\documentclass[10pt,twocolumn,letterpaper]{article}

\usepackage{iccv}
\usepackage{times}
\usepackage{epsfig}
\usepackage{graphicx}
\usepackage{amsmath}
\usepackage{amssymb}
\usepackage{comment}
\usepackage[accsupp]{axessibility}  % Improves PDF readability for those with disabilities.

\DeclareMathOperator*{\argmin}{arg\,min}
\usepackage[pagebackref=true,breaklinks=true,colorlinks,bookmarks=false]{hyperref}

\iccvfinalcopy % *** Uncomment this line for the final submission

\def\iccvPaperID{10564} % *** Enter the ICCV Paper ID here

\ificcvfinal\fi

\begin{document}

%%%%%%%%% TITLE
\title{Structured Bird's-Eye-View Traffic Scene Understanding from Onboard Images }
% \title{Structured Traffic Scene Understanding from a Single Onboard Camera Image}
% \author{Yigit Baran Can\\
% CVL\\
% ETH Zurich\\
% {\tt\small yigit.can@vision.ee.ethz.ch}
% \and
% Alexander Liniger \\
% Computer Vision Lab \\
% ETH Zurich \\
% %   CVL - ETH Zurich\\
% {\tt\small alex.liniger@vision.ee.ethz.ch} \\
% \and
% Danda Paudel \\
% Computer Vision Lab \\
% ETH Zurich \\
% {\tt\small paudel@vision.ee.ethz.ch} \\
% \and
% Luc Van Gool \\
% Computer Vision Lab \\
% ETH Zurich \\
% {\tt\small vangool@vision.ee.ethz.ch}
% \author{
% Yigit Baran Can \and Alexander Liniger\\
% CVL - ETH Zurich \and Leuve\\
% }
\author{ Yigit Baran Can\textsuperscript{1}\space\space\space\space Alexander Liniger\textsuperscript{1}\space\space\space\space Danda Pani Paudel\textsuperscript{1}\space\space\space\space Luc Van Gool\textsuperscript{1,2}\\
\textsuperscript{1}Computer Vision Lab, ETH Zurich\space\space\space\space \textsuperscript{2}VISICS, ESAT/PSI, KU Leuven \\ {\tt\small $\{$yigit.can, alex.liniger, paudel, vangool$\}$@vision.ee.ethz.ch} }
\maketitle
\ificcvfinal\thispagestyle{empty}\fi

%%%%%%%%% ABSTRACT
\begin{abstract}
Autonomous navigation requires structured representation of the road network and instance-wise identification of the other traffic agents. Since the traffic scene is defined on the ground plane, this corresponds to scene understanding in the bird's-eye-view (BEV). However, the onboard cameras of autonomous cars are customarily mounted horizontally for a better view of the surrounding, making this task very challenging. In this work, we study the problem of extracting a directed graph representing the local road network in BEV coordinates, from a single onboard camera image. Moreover, we show that the method can be extended to detect dynamic objects on the BEV plane. The semantics, locations, and orientations of the detected objects together with the road graph facilitates a comprehensive understanding of the scene. Such understanding becomes fundamental for the downstream tasks, such as path planning and navigation. We validate our approach against powerful baselines and show that our network achieves superior performance. We also demonstrate the effects of various design choices through ablation studies. Code: \url{https://github.com/ybarancan/STSU}
\end{abstract}

\section{Introduction}

\begin{figure}
    \centering
    \includegraphics[width=\linewidth]{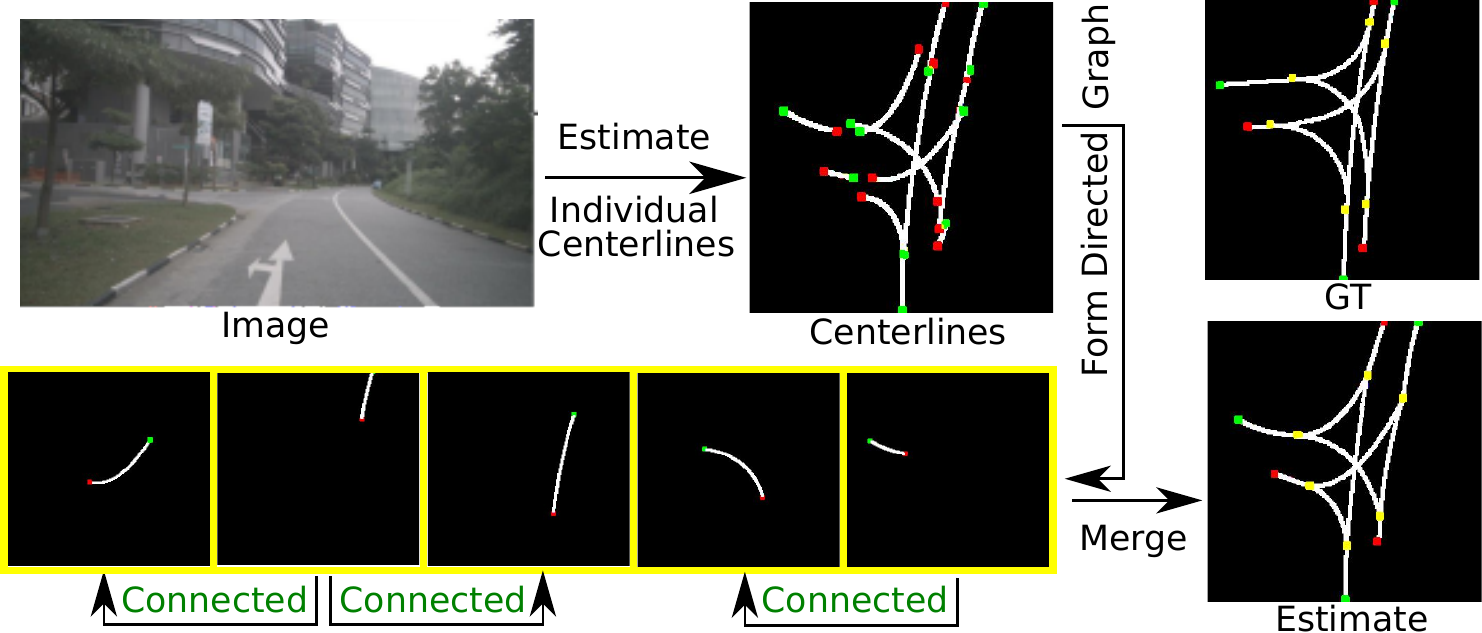}
    % \caption{Our method extracts a directed graph that represents the local road network,  using a single image. Green dots indicate the start point and red dots indicate end point of the centerline. Traffic flows from green to red. This representation can be achieved thanks to directional nature of the proposed method.}
        \caption{Our method extracts a directed graph that represents the local road network from a single frontal image. First, centerlines are estimated alongside a directed graph where vertices are centerlines, and edges show the connectivity. Then the existence and direction of the edges are estimates. Green dots indicate start points, and red dots indicate end points of centerlines. Traffic flows from green to red. This representation can be achieved thanks to the directional nature of the proposed method.}
    \label{fig:teaser}
    \vspace{-1em}
\end{figure}

Road scene understanding is crucial for autonomous driving since it forms the interface between perception and planning. The fundamental task is to understand both the road network structure and the other traffic agents in the surrounding. Currently, the go-to solution is offline generated HD-maps combined with a modular perception stack~\cite{jaritz20202d,seif2016autonomous,ma2019exploiting,ravi2018real,casas2021mp3}. For existing solutions to work, not only the precise localization in the HD-map but also understanding the dynamic parts of the scene is necessary~\cite{ma2019exploiting,yang2018hdnet}. To achieve these requirements, most solutions use several sensors, including cameras and LIDAR. However, using expensive sensors and offline HD-maps limit the scalability of autonomous driving as they increase the cost of operation and limit self-driving cars to operate in geographically restricted areas.

\begin{figure}
    \centering
    \includegraphics[width=.9\linewidth]{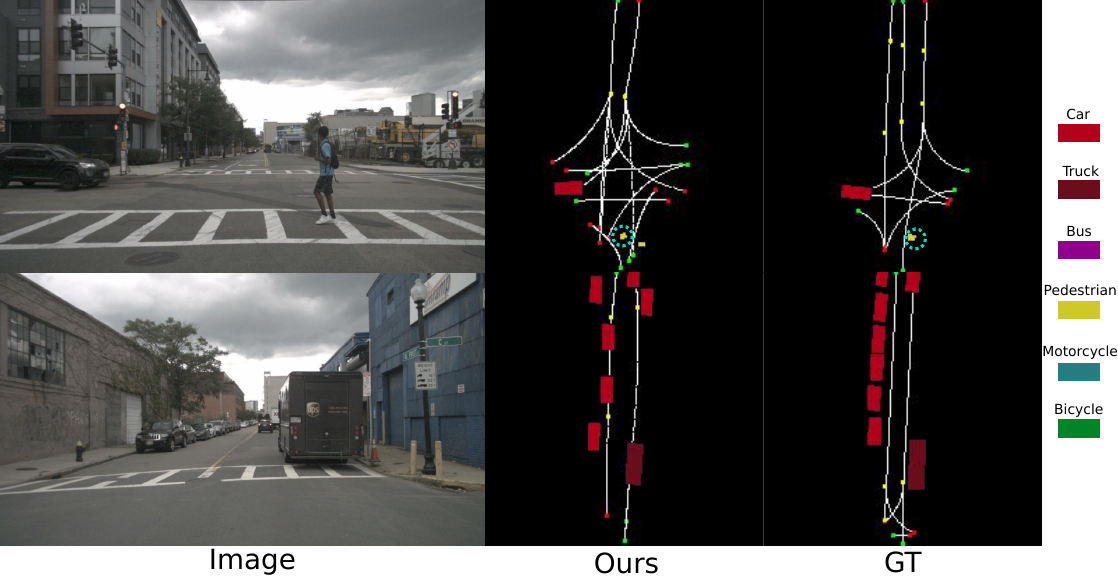}
    \caption{Our method can handle very complex cross roads scenes as well multiple object instances. Pedestrian is marked with circle.}
    \label{fig:teaser2}
    \vspace{-1em}
\end{figure}

In this work we proposes an end-to-end vision method that performs lane-graph extraction and object detection given only a front-facing camera image. Our method directly estimates the graph structure of the road network and spline curves representing centerlines of individual lanes, as shown in Fig.~\ref{fig:teaser}. Besides estimating the road graph, our model can also detect objects such as cars, pedestrians, and others, directly on the BEV plane, as shown in Fig.~\ref{fig:teaser2}. The output format of our method is ideal for downstream planning~\cite{DBLP:conf/rss/BansalKO19,chen2020learning} and prediction~\cite{cui2019multimodal,zaech2020action,hong2019rules,rella2021decoder} tasks, which require both the lane-graph and the location and class of objects. In fact, such a requirement can also be understood simply by observing the provided labels of existing datasets, such as~\cite{nuscenes2019}, which provide the labels in a structured form. Often, existing approaches map the structured labels into other forms, such as semantic masks, to perform scene understanding~\cite{garg2021semantics}. The downstream tasks, however, require the structured form of these understandings~\cite{liang2018end,paz2021tridentnet,liang2019convolutional,homayounfar2019dagmapper}.

Understanding HD-maps is a challenging problem, mainly due to the complex topological changes. Recovering such topological structure coherently from a single image remains to be an unexplored problem. This work addresses this challenging problem for the first time while also detecting objects in the scene directly in the BEV coordinates. Existing works either focus on (i) HD-map extraction from dense 3D points~\cite{homayounfar2019dagmapper} or (ii) the detection of road lanes from a single image~\cite{hou2019learning}. Other variants, such as BEV semantic understanding, also exist~\cite{DBLP:conf/cvpr/RoddickC20,DBLP:journals/ral/LuMD19,DBLP:journals/ral/PanSLAZ20}. Note that the HD-map reconstruction of~\cite{homayounfar2019dagmapper} is much more topologically challenging than the lane detection problem of~\cite{hou2019learning}. Our work aims to achieve results similar to~\cite{homayounfar2019dagmapper} using the image input setup of~\cite{hou2019learning}. Additionally, we aim to detect objects using the same model as for structured HD-map predictions.

We represent the HD-map as a directed graph in BEV coordinates, whose edges are the road segments and the direction represents the traffic flow. We model each road segment using a Bezier curve, with starting and end points. The connections between the predicted segments are modeled using an assignment matrix. For the prediction, we make use of a transformer network, which is supervised by using the Hungarian algorithm at the output end. The predicted segments, along with their connectivity, defines a full lane graph HD-map. Our transformer network further predicts the parameters of 2D BEV objects. The object prediction branch is supervised, similar to the road segments. Two example outputs of our method for both lane graph HD-map and object estimation are shown in Fig.~\ref{fig:teaser2}. To this end, our major contributions can be summarized as follows. 
% \vspace{-0.2cm}
\begin{itemize}
\setlength{\itemsep}{0pt}
\setlength{\parskip}{0pt}
\item We propose a unified method for structured BEV road network graph estimation and object detection from a single onboard monocular camera image.
% \item A novel and generic network architecture for structured output that can potentially be used beyond the addressed problem has been proposed in this paper. 
\item The results obtained by the proposed method are significantly superior to the compared baselines.
\end{itemize}

\section{Related Works}

\noindent\textbf{Road network extraction:} Early works on road network extraction use aerial images~\cite{auclair1999survey, richards1999remote}. Building upon the same setup, recent works~\cite{batra2019improved,sun2019leveraging,ventura2018iterative} perform the network extraction more effectively. However, aerial imaging-based approaches only provide coarse road networks. Such predictions may be useful for routing, however, they are not accurate enough for action planning. 

\noindent\textbf{High definition maps:} In the literature, HD-maps are often reconstructed offline using aggregated 2D and 3D visual information~\cite{liang2019convolutional, homayounfar2018hierarchical,liang2018end}. Although these works are the prime motivation behind our work, they require 3D point clouds for accurate HD-map reconstruction. More importantly, the offline methods recover the HD-maps in some canonical frame. Thus, using the recovered maps requires accurate localization, in many cases. In this regard, our work is similar to~\cite{DBLP:conf/cvpr/HomayounfarMLU18}, where the lane boundaries are detected on highways in the form of polylines. An extension of this work~\cite{DBLP:conf/cvpr/HomayounfarMLU18} uses a recurrent neural network to generate initial boundary points in 3D point clouds. The initial points are then used as seeds for a Polygon-RNN~\cite{DBLP:conf/cvpr/AcunaLKF18} that predicts lane boundaries. Our method differs from~\cite{DBLP:conf/cvpr/HomayounfarMLU18} in two major aspects: (i) point clouds vs. single image input, (ii) highway lane boundaries vs. lane centerlines in an unrestricted setting.

\noindent\textbf{BEV semantics understanding:}
Because of its practical use, scene understanding in BEV using images has recently gathered significant attention~\cite{DBLP:conf/cvpr/RoddickC20,philion2020lift,can2020understanding}. Some methods also combine images with LIDAR data~\cite{pan2020cross,hendy2020fishing}. In this regard, methods developed in~\cite{DBLP:conf/cvpr/RoddickC20,DBLP:journals/corr/abs-2002-08394} use a single image to understand the BEV HD-map semantics. Similarly, the method proposed in~\cite{can2020understanding} uses video data for the same task. Methods in this category do not offer structured output suitable for many downstream tasks. These methods may be used for general scene understanding. However, their usage for the task of motion planning and navigation is not trivial. Furthermore, up to our knowledge, no existing method provides instance-level predictions on the BEV using single image input. Note that the method proposed in this paper predicts both HD-map and the road objects simultaneously, using one input image and a single neural network.   

\section{The Proposed Method}
The core task of our model is to produce a directed graph that represents the road network in a BEV coordinate system, given only a single image from a front-facing camera mounted on a vehicle. For the complete traffic scene understanding, our model also outputs  objects' instances in the form of BEV bounding boxes. Both these tasks require reasoning about the 3D space and projecting all the information on the BEV ground plane where the vehicle is moving.

In this section, we first introduce our trainable lane graph structure and describe the object representation. Given these building blocks, we introduce our transformer based model and explain how the neural network is trained. 
% The proposed method takes a single front camera image as input and produces a directed graph that represents the road network in a BEV coordinate system for the visible region. 
% Apart from estimating the road network, it also outputs instance detections in the form of 2D bounding boxes that represent the projections of the objects in the BEV coordinate system. 

% \subsection{Road Scene understanding with Transformers}
% What is the correct order?
% \begin{itemize}
%     \item 3.0 rough outline
%     \item 3.1 lane graph representation (centerline + graph)
%     \item 3.2 object representation
%     \item 3.3 Transformer Architecture (much more detailed than it is now)
%     \item 3.3.1 general info (query vectors, positional encoding, ....)
%     \item 3.3.2 lane graph branch + association module
%     \item 3.3.2 object branch
%     \item 3.4 Training
    
%     \item 4 Metrics and Evaluation (it is a new task we can discuss this quite extensively)
    
% \end{itemize}

% \begin{itemize}
%     \item 3.0 rough outline
%     \item 3.1 Transformer Architecture
%     \item 3.2.1 lane graph representation (centerline + graph)
%     \item 3.2.2 training lane graph
%     \item 3.3.1 object representation
%     \item 3.3.2 training objects
%     \item 3.3  (much more detailed than it is now)
% \end{itemize}

\begin{figure*}
    \centering
    \includegraphics[width=1\linewidth]{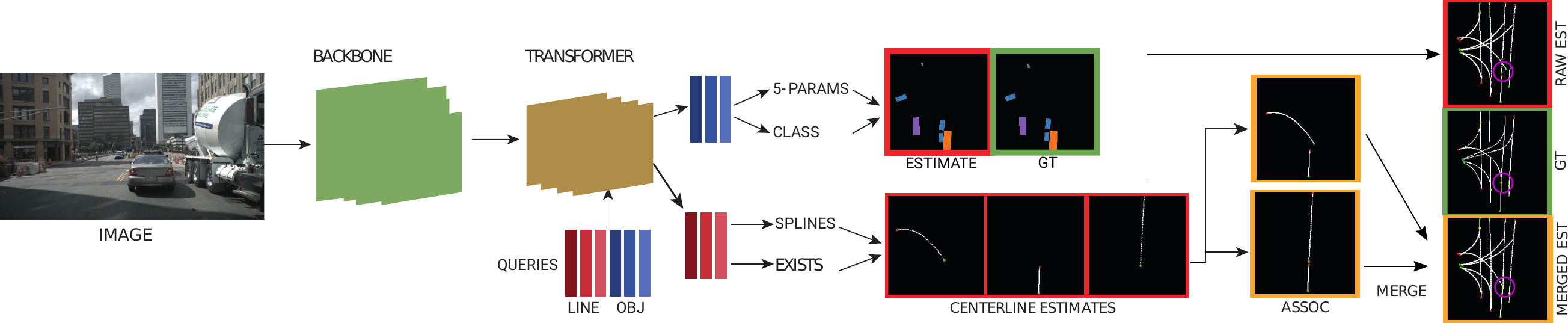}
    \caption{The core architecture of our neural network is a transformer \cite{DBLP:conf/eccv/CarionMSUKZ20} that processes learned centerline and object queries together. The processed line queries are used to output detection probability, control points, and centerline association features. The object queries are used to calculate the class probability and the oriented box parameters.}
    \label{fig:temp_net}
    \vspace{-1em}
\end{figure*}

\subsection{Lane graph representation}
In order to have a structured representation of the local road network, we build a directed graph of lane centerline segments, often called the lane graph. Let this directed graph be $G = (V, E)$ where $V$ are the vertices of the graph (the centerlines) and the edges $E \subseteq \{(x,y)\; |\; (x,y) \in V^2\}$ represent the connectivity among those centerlines. The connectivity can be summarized by the incidence matrix $I$ of the graph $G = (V, E)$. Thus, let us define when two centerlines $x$ and $y$ are connected; A centerline $x$ is connected to another centerline $y$, i.e. $(x,y) \in E$ if and only if the centerline $y$'s starting point is the same as the end point of the centerline $x$. Given this definition, an entry of the incidence matrix $I[x,y] = 1$ if the centerlines $x$ and $y$ are connected. Note that we do not apply a hard requirement to generate acyclical graphs, but cycles rarely occur due to our focus on a single image with a limited field-of-view (FOV). Thus, the incidence matrix often has the structure of an acyclical graph, where the main diagonal is zero and the sum of symmetric entries is at most one. % This indicates that the incidence matrix most of the time has a $\textbf{0}$ main diagonal and the sum of symmetric entries can be at most 1.
Finally, the resulting incidence matrix also contains crucial information about the traffic flow directions, which is fundamental to understand a lane network.

With the graph established, we need to model each centerline (vertex of the graph) mathematically. In this work, we consider each centerline as a Bezier curve. A Bezier curve maps a scalar parameter $t \in [0,1]$ to a point in $\!\mathbb{R}^\Delta$. We are interested in 2D curves for our lane graph, thus $\Delta = 2$. The curve can be written as the weighted sum of control points $\!P = \{P_0, P_1, ..., P_n\}$ where $P_i \in \!\mathbb{R}^2$. Given the control points, the curve $B$ parameterized by $t$ is defined as $B(t) = \sum_{k=0}^n \binom{n}{k} (1-t)^{n-k}t^k P_k$. A more compact matrix-based formulation is simply $B(t) = \boldsymbol{\Gamma}(t,n)\boldsymbol{P}$, where $\boldsymbol{\Gamma}_{ij}(t,n) =\binom{n}{j} (1-t_i)^{n-j}t_i^j$ represents the weight matrix, and $\boldsymbol{P}=[P_0, P_1,...,P_n]$ is the vector of all control points. With this representation, finding the optimal control points given some observed points $\boldsymbol{Y}=[Y_0, Y_1,...,Y_T]$ amounts to solving a least square problem, i.e., $\boldsymbol{P^*}= \argmin_{\boldsymbol{P}} \|\boldsymbol{\Gamma}(t)\boldsymbol{P} - \boldsymbol{Y}\|$. 
Bezier curves are a good fit for centerlines since it allows us to model a curve of arbitrary length with a fixed number of 2D points.
% This formulation is very useful for our task since it allows us to model a curve of arbitrary length with a fixed number of 2D points. 
% Thus we have a learnable representation of a lane graph, where the network has to learn the centerlines in terms of Bezier control points, as well as the connectivity of the graph. 
Thus, given our graph and centerline representation, the whole lane graph has a fixed-sized learnable representation, where the network can learn the centerlines in terms of Bezier control points, and the connectivity of the graph.

\subsection{Object representation}
Our method also produces object instance detections to complement the lane graph and give a complete traffic scene understanding. Different than semantic segmentation, instance outputs localize and identify individual objects. We represent each instance as a 2D box in normalized BEV coordinates. In order to fully specify such a box, one needs five parameters: location of the center point, short and long side length, and the heading angle. 
% Since we do not differentiate between 180$^\circ$ flipped objects angles lay between $[0,\pi]$. 
From these parameters, it is a simple conversion to the four corner point locations and vice versa. Apart from the localization and orientation of the instances, we also produce their semantics/object class using a one hot representation. Given this representation, an instance is fully identified.

\subsection{Architecture}
We have modeled each centerline and object instance as a fixed size vector. Thus, we can work within the framework of proposal generation and classification. This has been widely used in the fields of instance segmentation and object detection \cite{DBLP:journals/pami/RenHG017, he2017, DBLP:conf/eccv/CarionMSUKZ20}. One crucial property of our formulation is the strong relationship among different centerlines as they form the lane graph. However, there is also a strong relation between centerlines and objects since, in traffic scenes, objects follow centerlines. In order to fully exploit this dependency, we adapt the transformer-based model proposed in \cite{DBLP:conf/eccv/CarionMSUKZ20}, which allows us to train one joint model for lane graph and object understanding. 

The transformer-based object detector proposed in \cite{DBLP:conf/eccv/CarionMSUKZ20} uses image backbone features and learnable query vectors to generate object proposals. We follow a similar approach, but we use two sets of learned query vectors $Q \in \!\mathbb{R}^{C}$: one set for centerlines and one for objects. The number of these vectors is higher than the maximum centerlines/objects that can occur in any scene. These query vectors are processed \emph{jointly} by the transformer, which outputs a proposal vector for each query. These vectors encode all the information needed to fully identify a centerline or object. Each of these proposal vectors is further processed to generate an output. This processing is done in a separate lane and object branch, which output the lane graph and object detections. The overall architecture is given in Fig.~\ref{fig:temp_net}. %The final outputs of the transformer are processed in a separate lane and object branch.

\subsubsection{Lane branch}
The first of the two branches processing the output vectors of the transformer is the lane branch, which has four parts:
% The lane branch consists out of four main parts:

\noindent\textbf{Detection head:} The transformer output is processed by a multi-layer perceptron (MLP) with an output softmax layer. This output gives the probability that the centerline encoded by the corresponding query vector exists. 

\noindent\textbf{Control head:} An MLP + sigmoid layer with $2\times R$ output neurons, encoding the $R$ Bezier curve control points.

\noindent\textbf{Association head:} An MLP that outputs a $\delta$-dimensional association feature vector for each of the centerline vectors, where $\delta < C$. The classifier uses these association features to establish the connectivity of the estimated centerlines.   

\noindent\textbf{Association classifier:} An MLP + sigmoid layer, which takes two $\delta$ dimensional association feature vectors corresponding to two centerlines as an input. This layer outputs the probability of the input centerline pair being associated.   

As a first step of estimating the graph, we extract individual centerlines. This is done by the detection and control heads. These lines form the vertices $V$ of the graph $G = (V, E)$. Given that $N$ centerlines are selected, the feature vectors of the corresponding centerlines are processed by the association head to produce lower-dimensional association feature vectors $F \in  \!R^{N \times \delta}$. Then, we obtain the association inputs $A \in \!\mathbb{R}^{N \times N \times 2\delta}$, where $A_{ij} = \text{concat}([F_i, F_j]) $. This input encodes the directional nature of the graph. An MLP processes the matrix $A$ to produce the incidence matrix probabilities. Note that the MLP has an input dimension of $2\delta$, and $N \times N$ is the batch size. Thus there is no constraint on the number of proposed centerlines.
% To further illustrate the association classification module, we give a visual representation of the task in Fig.~\ref{fig:assoc_exp}, where the association between four centerlines is visualized. The asymmetric pattern in the positive associations, highlighted in green, shows the how this problem also considers traffic flow. 
% An example of association classification is given in Fig~\ref{fig:assoc_exp}, where the asymmetric nature of this process is shown.

During training, we first output centerline control points and detection probabilities and apply the Hungarian matching algorithm among the estimated  and the ground truth (GT) centerlines. The association step is carried out on the matched estimates. During inference, we threshold the detection probability of the centerlines and carry out the association step on the active lines.

\subsubsection{Object branch}
The second branch that processes the transformer proposal vectors is the object branch. The branch consists of two modules and an optional post-processing network.

\noindent\textbf{Detection head:} The transformer output is processed by an MLP with a softmax output layer to produce class probability distribution, including a ``no detection" class.  

\noindent\textbf{5-params head:} An MLP + sigmoid layer that produces the normalized parameters of the oriented object boxes.% In order to get the angle, we multiply the angle output by $\pi$. 

\noindent\textbf{Refinement net:} While the instance outputs are suitable for many tasks, it is also beneficial to produce semantic segmentation maps of the scene. This is especially true for small objects like pedestrians, and bikes, where the localization in the BEV from a single onboard camera is difficult. Therefore, we propose an optional post-processing network that converts the instance estimations to semantic segmentations using our refinement net. The structure of the refinement net is similar to the BEV decoder of \cite{DBLP:journals/corr/abs-2012-03040}, where a lower resolution input is upscaled to provide a fine-grained segmentation map. The network operates as follows: there are $C+1$ classes including the background, and the region of interest is $H \times W$ dimensional. We first convert the 5-params output to a box and multiply this box with the class probabilities. This results in a matrix $M$ of dimension $H \times W \times (C+1)$ where $\sum_i M_{h,w,i} = 1$ for all grid locations $(h,w)$ that fall in the bounding box and 0 otherwise. Then we sum all these matrices and clip them to (0,1). The result is again of dimension $H \times W \times (C+1)$. To inform refinement net about the visual cues in the image, we also include backbone features. Since the bounding box locations are in the BEV coordinates, we warp the backbone features to the BEV and use warped skip connections in the upsampling.% Additionally the refinement net also takes warped backbone features and backbone skip connections as input. 

% \begin{figure}
%     \centering
%     \includegraphics[width=.8\linewidth]{figures/assoc_exp.pdf}
%     \caption{The process of association estimation is shown. Green borders mean that the estimation pair is assigned high association probability by the classifier while red is otherwise. It can be seen that the lines in rows are the initial lines and the lines in column should be following them. }
%     \label{fig:assoc_exp}
%     \vspace{-1em}
% \end{figure}

\subsubsection{BEV positional embedding}
Since transformers do not have a notion of position, positional embeddings (PE) are used to add spatial awareness. We use two different positional embeddings for the transformer. The first one encodes the image domain spatial information where similar to \cite{DBLP:conf/eccv/CarionMSUKZ20}, we use sinusoidal functions on the normalized cumulative locations. The second set of positional embeddings encode the corresponding BEV location of a given pixel. For this purpose, we assume a flat surface where every real world-point has height $-C_h$, where $C_h$ is the camera height. The resulting grid is very dense in image coordinates that correspond to real-world points close to the camera while it is sparse for further away positions. In order to provide more uniform location cues to the network, we use the logarithm of BEV locations. To generate the positional encoding, we take the cumulative sum, normalize, and convert it into a sinusoidal. We design the two positional embeddings (image and BEV) such that they are half of the channel size of the input feature map. Thus, we can add the image positional embeddings to half of the channels while BEV positional embeddings are added to the other half. The main reason for this design choice is that BEV coordinates are undefined for the upper half of the image, but they still hold important cues for the network about objects in this region. Simply adding these two positional embeddings would result in duplicates and inconsistent values in the lower half of the image. We dubbed this use of dual positional embeddings as \emph{split positional encoding}.

\subsubsection{Training}
Since both the lane and object branches produce detection-like outputs, we use Hungarian matching on the estimations and the ground truth during training. The matching loss used for lanes and objects is similar. Both are of the form $L_{m} = L_{CE} + \lambda L_1$, where $L_{CE}$ is the cross-entropy loss on the detection/class probability, and $L_1$ is the 1-norm loss on the Bezier control point locations/Box parameters for lines and objects, respectively. For both the centerline and object branch, the training detection loss is cross-entropy.
Control point and object 5-params outputs are trained using $L_1$ loss, except for the angle of objects. Since 180$^\circ$ object flips are hard to distinguish, we predict only angles in the range $\alpha \in [0,\pi]$ and train it using a smooth sin/cos $L_1$ loss of the form $L_{angle}=|\text{cos}(2\alpha) - \text{cos}(2\phi)| + |\text{sin}(2\alpha) - \text{sin}(2\phi)|$, where $\phi$ is the GT angle. We train the refinement net independently using the cross-entropy loss, not backpropagating through the rest of the network. 
\section{Metrics}

Since our problem setup is new, there is a lack of suitable metrics for performance evaluation. We wish to measure the performance in reproducing the real directed graph faithfully. For this purpose, we use three metrics that aim to highlight different aspects of the directed graph.

\begin{figure}
    \centering
    \includegraphics[width=.7\linewidth]{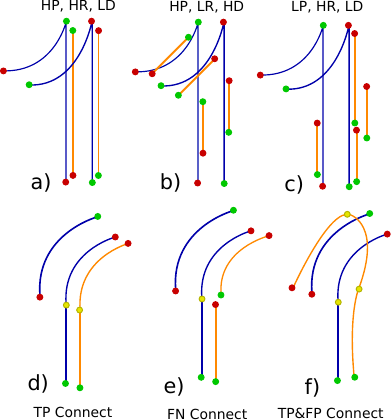}
    \vspace{0.2cm}
    \caption{Some examples of precision-recall \& detection (a,b,c) and connectivity metric (d,e,f). Blue lines are true centerlines and orange are estimations. Green/red dots represent starting and end points, respectively. Yellow points indicate a connection, which is only valid in the direction from green to red. H/L refers to High/Low while P=Precision, R=Recall and D=Detection. a) 2 out of 4 lines are missed but the matched lines are accurate. b) Matched true lines are longer than estimates creating false negatives. c) All estimates are matched to one true line (note the endpoint colors on the leftmost estimate), leaving no room for false negatives while creating false positives. d) and e) Show true positive and false negative connectivity, respectively. f) One connection is a true positive but the upmost connection is a false positive.}
    \label{fig:metric_fig}
    % \vspace{-1em}
\end{figure}

\subsection{Precision-Recall}
Following \cite{DBLP:journals/corr/abs-2012-12160, DBLP:conf/cvpr/HomayounfarMLU18}, we calculate precision-recall on matched centerlines at different distance thresholds. We first match each estimation to the target with the minimum $L_1$ loss on Bezier coefficients. Thus, similar to \cite{DBLP:journals/corr/abs-2012-12160}, multiple estimations can be matched with the same target while each estimation can only be matched with one target. Then we interpolate the estimated coefficients to get dense centerlines. Note that using control points during matching is fundamentally different from using interpolated points because the control point based matching takes direction into account. Thus, two centerlines where only the order of control points is reversed (start and end points are swapped) are identical if interpolated points are matched. However, they are far apart in our control point matching approach. After matching based on the control points and then interpolation, a true positive is an estimated interpolation point within a threshold distance to the matched GT line and a false positive otherwise. A false negative is a point on a GT line that is not within the threshold distance of any of the matched estimated lines. Note that this metric does not penalize the missed centerlines, i.e., true centerlines that are not matched with any estimation. This is intentional since the focus of this metric is measuring how well the estimates fit the matched GT centerlines and how accurately the captured subgraph is represented.   

\subsection{Detection ratio}

In order to measure the aforementioned issue of missed centerlines present in the precision-recall metric, we calculate the detection ratio. This is simply the number of unique GT centerlines that at least one estimated line is matched to over the total number of GT centerlines. High scores in precision-recall and a low detection score mean that the estimated centerlines are close to the matched true ones, but a substantial part of the GT centerlines is not detected. The inverse implies that the estimated centerlines cover the true road network but do not faithfully represent the structure. These two metrics summarize the performance on the vertices of the total graph $G$. However, we still lack a metric to evaluate performance on the edges of the graph, i.e., the connectivity.

% \begin{figure}
%     \centering
%     \includegraphics[width=.7\linewidth]{figures/metric_figure.pdf}
%     \caption{a) Estimations produce low false positive and false negatives with their matched targets but a sizable portion of the road network is missed. b) Estimated (marked) centerlines are shorter than their matched true line and produce false negatives dropping recall while there is not many false positives. The leftmost line is missed reducing detection. c) While all the true lines are found, extra estimations are producing false positives, reducing precision. }
%     \label{fig:metric_fig}
%     \vspace{-1em}
% \end{figure}

% Apart from precision-recall, we have produced an mIOU based metric. In order to apply mIOU, we generate polygons from the estimated spline coefficients. For each pair of subsequent spline coefficients, we obtain a line segment. then, given a lane width, half of that lane width is set as the magnitude of the vectors perpendicular to the obtained lane segment. These vectors are added to both spline coefficients to obtain 4 polygon points. When all the pairs are traversed, the resulting polygon represents the lane. The same procedure is applied to the true centerlines. Mean IOU between these 2 give an overall idea about how much of the true road network is captured. 

\subsection{Connectivity}

In order to measure how well the estimated centerlines are associated, we propose a precision-recall-based metric, called connectivity metric. Let the estimated binary incidence matrix be $E$, and the GT incidence matrix be $I$. Let $M(i)$ be the index of the target that the $i$th estimation is matched to and $S(n)$ be the set of indices of estimations that are matched to target $n$. A positive entry $E_{ij}$ is a true positive if ${(M(i) == M(j)) \;|\; (I(M(i), M(j)) == 1)}$, and a false positive otherwise. On the other hand, a false negative is a positive entry of the incidence matrix $I_{m,n}$ where ${\nexists \;(i,j) : ((i \in S(m))\; \& \;(j \in S(n))\; \& \;(E_{i,j}==1))}$. 

% In words, if two estimated centerlines are associated, there are two possible ways for this association to be true:
% \vspace{-.5em}
% \begin{itemize}
% %  \itemsep-.2em
% %  \setlength\itemsep{0em}
% \itemsep-.4em 
%     \item Both estimations are matched with the same target
%     \item The distinct targets that the two estimated centerlines are matched to are, indeed, associated according to the GT incidence matrix
% \end{itemize}
% \vspace{-.5em}
% A miss, or a false negative, is when there is a positive entry $(m,n)$ in the GT incidence matrix $I$ and at least one of the following conditions hold: 
% \vspace{-.5em}
% \begin{itemize}
%  \itemsep-.4em 
%     \item No estimation was matched with target $m$
%     \item No estimation was matched with target $n$
%     \item Among all pairs of estimated centerlines $(i,j)$ where $i$ is matched with target $m$ and $j$ with $n$, there is no pair whose association estimate is positive
% \end{itemize}
% \vspace{-.5em}
This metric captures how close the connectivity pattern of the estimated graph is to the GT graph. With this metric, fragmenting a true centerline into multiple estimations is not a problem as long as they are associated. Some graphical illustrations of the three metrics are given in Fig.~\ref{fig:metric_fig}.

\section{Experiments}

\subsection{Dataset}

We use the NuScenes \cite{nuscenes2019} dataset consisting of 1000 sequences recorded in Boston and Singapore. The sequences are annotated at 2Hz, and the dataset provides HD-Maps in the form of centerlines. The dataset also provides 3D bounding boxes of 23 object classes. For our experiments, we select the most frequent classes: car, truck, bus, pedestrian, bike, and motorcycle. We only use the front camera both for training and evaluations. 

Given the set of real-world coordinates of a particular centerline, we first convert these coordinates to the camera coordinate system of the current reference frame. We resample these points with the target BEV map resolution and discard any point outside the region-of-interest (ROI). The points are then converted to normalized coordinates given the bounds of the ROI. This results in a set of points between [0,1], from which we extract the control points of the Bezier curve. The ground truth labels and the estimates are in normalized control point coordinates. 

\subsection{Implementation}
We use images of size 448x800, and the target BEV area extends from -25 to 25m in x-direction and 1 to 50m in z-direction. BEV resolution is set to 25cm. Given the common structure of roads, the possible complexity of the curves that represent the centerlines segments is limited. Thus, we use three Bezier control points. We use two sets of 100 query vectors for centerlines and objects: one for right (Boston) and one for left-sided traffic (Singapore). The backbone network is Deeplab v3+ \cite{DBLP:conf/eccv/ChenZPSA18} pretrained on Cityscapes dataset \cite{Cordts2016Cityscapes}. The implementation is in Pytorch. The method runs with 11FPS without batching and including all association and refinement steps.

% We apply artificial depthwise motion during training as a data augmentation. With the flat world assumption, it is possible to calculate the new pixel location of a real world point if the ego vehicle moves $\beta$ in depth direction. Let the original pixel row and column coordinates be $(m_0, n_0)$ and the new coordinates be $(m_1, n_1)$. Then $n_0 = (n_1-d_x)fC/(fC - m_1\beta + d_y\beta) + d_x$ and $m_0 = (m_1-d_y)fC/(fC - m_1\beta + d_y\beta) + d_x$ where $f$ is the focal length, $C$ is the camera height and $(d_x,d_y)$ are frame center coordinates. We resample the original image and translate object and centerline points by $\beta$. 

\begin{figure}
    \centering
    \includegraphics[width=\linewidth]{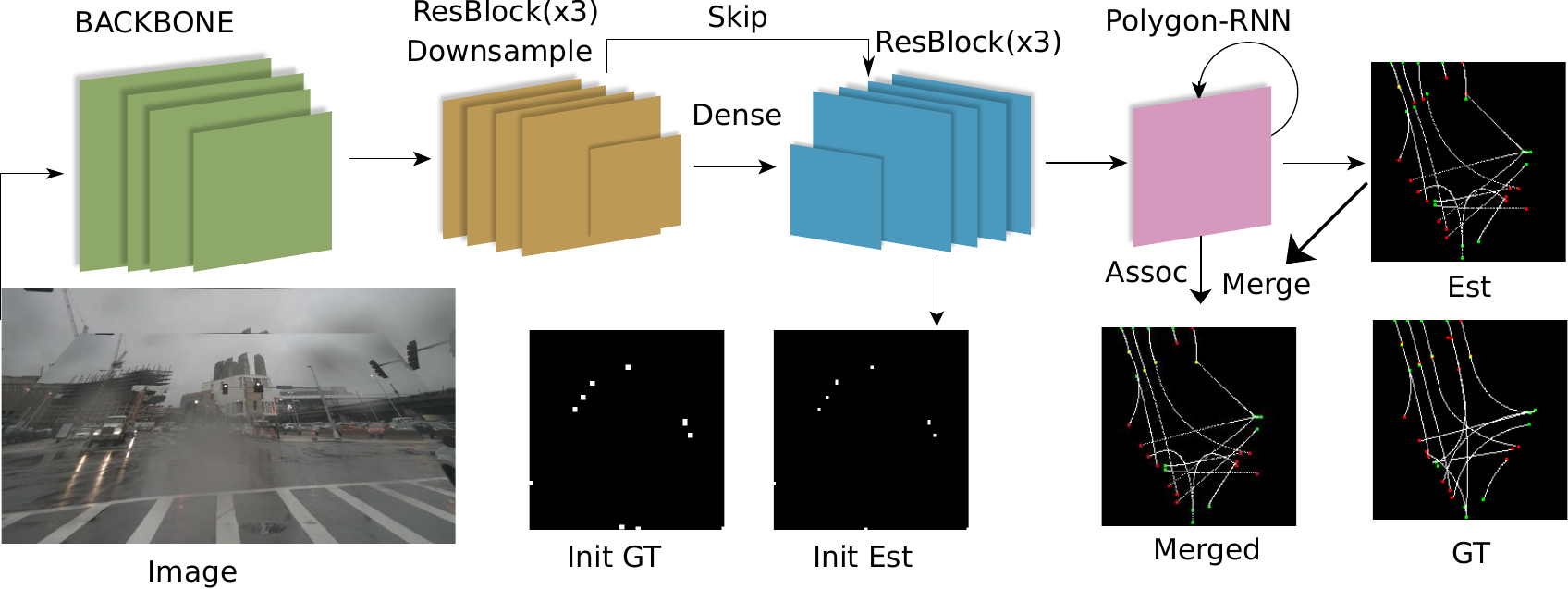}
    \caption{Polyline-RNN based method first extracts initial point estimations. Polygon-RNN uses the backbone features and initial point estimations to form the centerline curves. }
    \label{fig:polygon_rnn}
    % \vspace{-1em}
\end{figure}

\begin{figure*}
    \centering
    \includegraphics[width=0.85\linewidth]{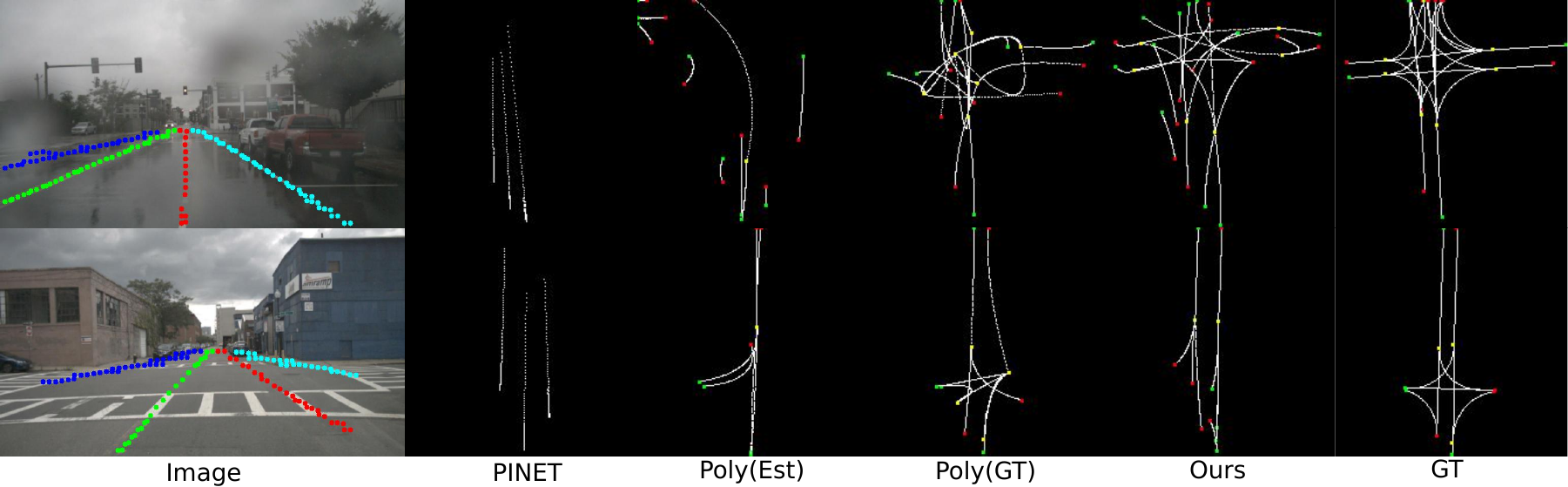}
    \caption{Sample centerline estimates. PINET boundary estimations are shown on the image. Our method produces the best lane graph representation. Statistical results for each sample are provided in the supplementary material. }
    \label{fig:lane-visual}
    % \vspace{-1em}
\end{figure*}
\subsection{Baselines}
Since there does not exist any method that deals with structured BEV road network estimation from a monocular image, we have generated two baselines. The first baseline is based on \cite{DBLP:conf/cvpr/HomayounfarMLU18}, where the authors generate lane boundaries from point clouds. We adapt their method to work with images and to output centerlines rather than lane boundaries. To achieve this, we project Deeplabv3+ \cite{DBLP:conf/eccv/ChenZPSA18} backbone features of the image to the BEV with a GT projection matrix. We concatenate x-y grid locations with this backbone feature map, similar to \cite{DBLP:conf/nips/LiuLMSFSY18}. A subnetwork with a fully connected layer at its core takes this input and outputs a grid of $49\times50$ points. 
We tested a convolutional RNN similar to the original work but did not achieve satisfactory results in our setting due to the restricted FOV. Note that the original task of finding lane boundaries on highways from aggregated LIDAR scans is significantly different from finding initial centerline points in urban traffic scenes. Moreover, the RNN required prohibitively many iterations, especially considering that one frame contains more than 40 centerlines.  
Therefore, we used a fully connected layer supported by several residual blocks, see Fig. \ref{fig:polygon_rnn}. Given the initial locations and the backbone features, Polygon-RNN \cite{DBLP:conf/cvpr/AcunaLKF18} produces the next control points of the centerline. We fix the number of iterations of Polygon-RNN to the number of spline coefficients used to encode centerlines. We use the focal loss \cite{DBLP:journals/pami/LinGGHD20} for the initial point estimation and an $L_1$ loss to supervise the control point estimation of Polygon-RNN. In training, Polygon-RNN uses GT initial points similar to \cite{DBLP:conf/cvpr/HomayounfarMLU18}. For testing, we threshold the initial point estimations of the network and feed them to Polygon-RNN. To indicate the direction of traffic, we feed a binary indicator variable to the initial point estimator. The association estimation is done using the final feature map of Polygon-RNN using the same approach as in our network. For reference, we also report results with Polygon-RNN using GT initial points.

For our second baseline, we extract lane boundaries with the SOTA method of \cite{DBLP:journals/corr/abs-2002-06604}. The extracted lane boundaries are projected onto the BEV grid using the GT transformation. Given these lane boundaries, we form the closest pairs and compute the centerlines using splines. Since this method does not give us a direction, we use the predicted and a flipped centerline version during evaluation, meaning PINET matching is directionless. In a sense, PINET estimates are manually assigned correct directions.

% Then, for each point in a given reference boundary, we get the minimum distance to all other candidate boundaries. The candidate boundary with the minimum mean distance forms a couple with the reference boundary. Then, we fit splines to each of them and get the mid points of the spline control points. The resulting points serve as the spline control points of the estimated centerline. Since there is no way of extracting directional information from this method, in matching we flip the order of estimated control points and take the minimum of the two distances. Thus, PINET matching is directionless, making its task easier. 

For the evaluation of our method's instance estimations, we compare against VED \cite{DBLP:journals/ral/LuMD19}, VPN \cite{DBLP:journals/ral/PanSLAZ20} and PON \cite{DBLP:conf/cvpr/RoddickC20}. We use the same train/val split proposed in \cite{DBLP:conf/cvpr/RoddickC20} for \emph{both lane graph and object detection}. We also follow \cite{DBLP:conf/cvpr/RoddickC20} for the object label generation procedure. To compare results, we use the mIOU metric. Nevertheless, for future reference, we also present precision-recall results.   

% \subsection{Ablations}

% In order to assess the impact of different settings of our method, we carried out evaluations on NuScenes dataset. In these ablations we studied the effect of the size of the transformer and the different positional embeddings. 
\section{Results}

\begin{figure}
    \centering
    \includegraphics[width=\linewidth]{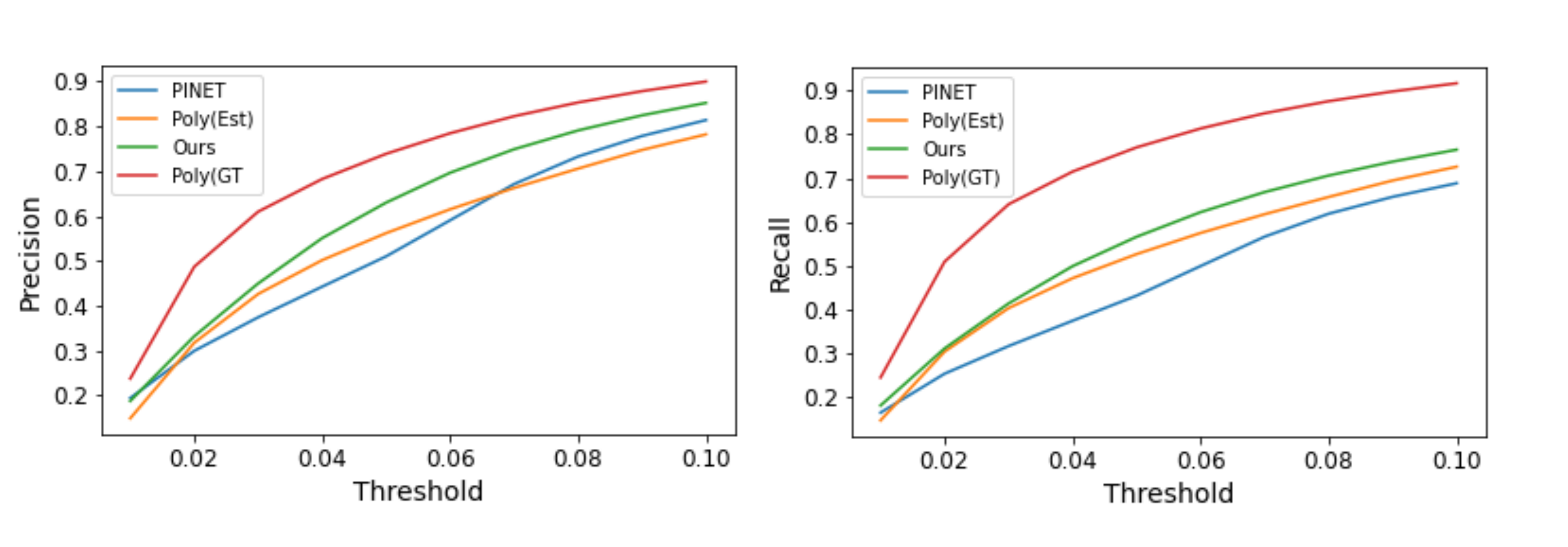}
    \caption{Precision/Recall vs thresholds. Thresholds are uniformly sampled in [0.01, 0.1] (normalized coordinates) with 0.01 increments. In our resolution, 0.01 corresponds to 50cm.}
    \label{fig:prec-recall}
    % \vspace{-1em}
\end{figure}
% \vspace{-1em}
Since our method produces a road network graph as well as dynamic object instance estimations, we divide the results into two subsections, studying those them individually.

\subsection{Lane graph}

The obtained results are provided in Tab.~\ref{tab:lane_compare} and Fig.~\ref{fig:prec-recall}, where our method achieves the best results in all metrics when compared to the baselines. The performance of PINET is lower, as expected, since the centerlines are obtained through processing lane boundaries. From the Poly(Est) vs Poly(GT) results, it can be seen that the localization of initial points is very difficult. Our method produces better precision-recall than Poly(Est), and the difference in detection and connectivity scores are significant. It is not surprising that Poly(Est) suffers in the connectivity metric, particularly connectivity recall. This metric is closely related to detection score, and missed centerlines are penalized. Our method's performance in connectivity precision against Poly(GT) combined with the detection scores shows that our method produces much fewer false-positive associations in the detected sub-graph and more accurately estimates the graph. The superiority of Poly(GT) in precision-recall and detection metrics is expected. Since most centerlines are relatively short and divergence from the initial point is limited, knowing GT initial points provides a clear advantage. However, its performance validates the strength of the  chosen baselines. 

\begin{table}[ht]
% \scriptsize{
\begin{center}
{
\tabcolsep=0.08cm
\begin{tabular}{ |c|c|c|c|c|c|c| }
\hline
Method & M-Pre & M-Rec & Detect & C-Pre & C-Rec & C-IOU  \\
\hline
PINET & 54.1 & 45.6 & 19.2 & - & -& -\\ 
Poly(Est) & 54.7 & 51.2 & 40.5 & 58.4 & 16.3 & 14.6\\
Ours& \textbf{60.7} & \textbf{54.7}& \textbf{60.6} & \textbf{60.5} & \textbf{52.2}&\textbf{38.9} \\
\hline
Poly(GT) & 70.0 & 72.3& 76.4 &53.8 & 52.0& 36.0\\
\hline
\end{tabular}
}
\end{center}
% }
% \vspace{-1em}
\caption{Lane graph results. M-Prec and M-Recall indicate mean of the sampled points of precision-threshold and recall-threshold curves, see Fig.~\ref{fig:prec-recall}. C-Prec and C-Rec refer to connectivity precision and recall, while C-IOU is connectivity TP/(TP + FP + FN).}
\label{tab:lane_compare}
\end{table}

\begin{figure*}
    \centering
    \includegraphics[width=.85\linewidth]{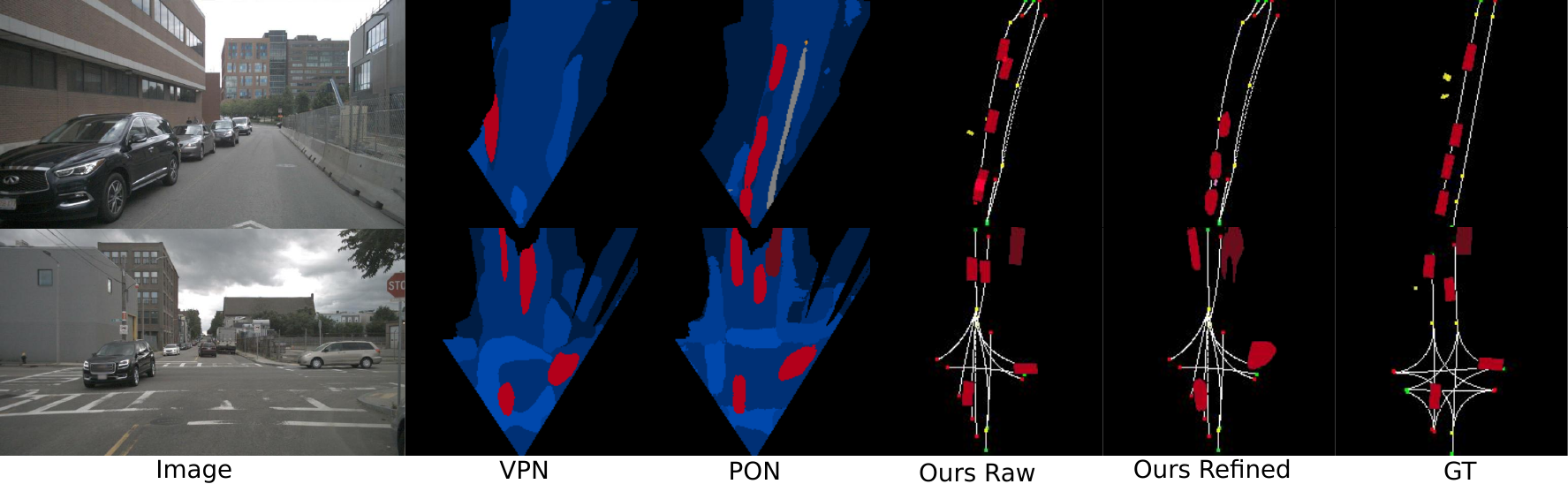}
    \caption{Visual results for object detection where we present the raw and refined estimates. We also show the road network estimates.}
    \label{fig:visual-obj}
    % \vspace{-1em}
\end{figure*}

Visual results for lane graphs are given in Fig.~\ref{fig:lane-visual}. Visual inspection shows that our method generally produces better results. In the last image, our method misses some centerlines. Overall, our method produces more faithful representations. On the other hand, Poly(GT) produces centerlines that are somewhat close, in the Euclidean sense, to the matched GT lines. However, the overall graph estimation is worse than ours. This shows the power of the connectivity metric where our method surpasses Poly(GT).

\begin{table*}[ht]

\begin{center}
{
\begin{tabular}{ |c|c|c|c|c|c|c|c|c|c|c|c|c|c|c|c| }
\hline
Method & M-pre & M-rec& detec & Con-IOU & car & truck & bus & ped & motor & bike & obj-mean\\
\hline
Large &  57.2 & 53.9& 58.8 & 41.0 & 20.0 & \textbf{11.7} &13.9 &  1.9 & 2.2 & 1.4 & 8.5\\
Large + Split&   59.9 & \textbf{56.8} & 52.8 & 40.8 & 20.0 & 10.1 & 16.8 &   1.9 & 2.8 & 0.8 &  8.7\\

Large + Split Log & \textbf{60.7} & 54.7 & 60.6 & 38.9 & 21.8 & 11.0 & 14.5 &  2.1 & 3.8 & 2.1 &  9.2\\

% Large + Split (big)& 51.4 & \textbf{34.0} & 14.8 & \textbf{21.9} & 12.8 & - & - & - & - & \textbf{20.4}& -\\

Small & 58.2 & 54.2 & 61.2 & \textbf{41.9} & 22.0 & 10.7 &15.1  & 2.0 & 2.9 & 1.7&  9.1 \\

Small+Split & 57.5 & 54.2 & 60.9 & 41.3 & 20.6 & 10.1 &14.0 &  2.0 & \textbf{4.1} & 2.3 & 8.9 \\
Small+Split Log & 58.9 & 53.6 & \textbf{61.5} & 37.8 & \textbf{22.6} & 10.9 & \textbf{17.6}  & \textbf{2.4} & 3.2 & \textbf{2.9}  & \textbf{9.9}\\
\hline
\end{tabular}
}
\end{center}
% \vspace{-1em}
\caption{Ablations are carried out on six models that test the performance contribution of the model size and positional embeddings. Object results are without refinement net and in the form of mIOU.}
\label{tab:ablation}
\end{table*}

\subsection{Objects}

In Tab.~\ref{tab:object_results}, the refinement net outputs of our network are compared against SOTA methods. Other methods usually produce estimates for slightly more classes. However, considering that we produce structured instance outputs along with lane graphs, we chose the most common yet comprehensive set of classes. Our method surpasses PON in half of the classes and in the mean measure. Especially, the difference in the ``car" category is rather significant.
\begin{table}[ht]

\begin{center}
\scriptsize{
\begin{tabular}{ |c|c|c|c|c|c|c|c|c|c|c|c|c|c| }
\hline
Method &  car & truck & bus &   ped & motor & bike & mean\\
\hline

VED &  8.8 & 0.2 & 0.0 &  0.0&  0.0&0.0&  1.5 \\ 
VPN &  25.5 & \textbf{17.3} & 20.0 &   7.1& 5.6 & 4.4 & 13.3 \\ 
PON &  24.7 & 16.8 &
20.8 &  \textbf{8.2}& 7.0 &\textbf{9.4} &14.5 \\ 
Ours& \textbf{32.5} & 15.7 & \textbf{21.2} & 6.2 & \textbf{7.4} & 6.4  & \textbf{14.9}\\
\hline
\end{tabular}
}
\end{center}
\vspace{-.5em}
\caption{Object results in mIOU of different methods.}
\label{tab:object_results}
\end{table}

The visual results for object estimates are given in Fig.~\ref{fig:visual-obj}. The competing methods  tend to blob segmentation and making harder to separate instances. Our refinement net outputs also suffer from the same phenomenon compared to our raw estimates. Despite of which, our refined estimates strike a good trade-off between mIOU maximization and instance separation.

\begin{figure}
    \centering
    \includegraphics[width=\linewidth]{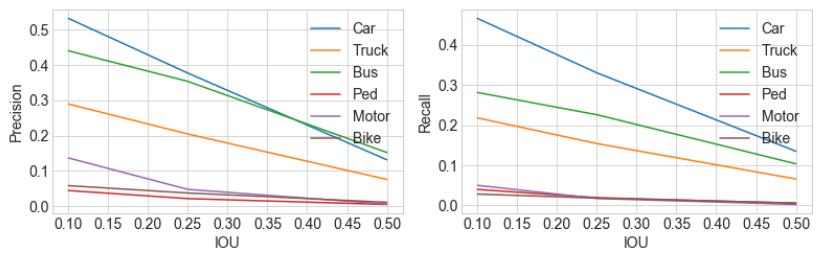}
    \caption{Precision/Recall vs IOU thresholds for object detection. We apply Hungarian matching with IOU to obtain corresponding estimate-GT pairs. If IOU is above the threshold, it is a true positive. Other GT objects count as false negatives, and the other estimates count as false positives.}
    \label{fig:obj-pre-recall}
    \vspace{-1em}
\end{figure}

\subsection{Ablation}

We experimented with two transformer sizes. The small model has two encoder layers and tree decoder layers, while the large one has four encoder and four decoder layers. We tried using vanilla positional embeddings and our split embedding with and without taking the logarithm. The results are given in Tab.~\ref{tab:ablation}, where the object results are in mIOU \emph{without} refinement net. We observe that our split embedding with log helps with objects, precision and detection scores while it causes a drop in connectivity IOU. Overall, the differences are rather low. Due to its good overall performance in object and lane results, we selected the ``Large+Split Log" model as the final one. When the object results of the selected model are compared with and without refinement net, the difference is rather significant. Refinement net boosts the performance by 5.7 points in mIOU.

\section{Conclusion}
We proposed a novel learnable representation of local road networks based on directed graphs and Bezier curve centerlines. This representation is used to train a transformer-based neural network architecture that predicts a complete lane graph structure from a single onboard image. We also proposed a set of metrics that are suitable to evaluate the performance of the proposed graph representation based structured scene understanding. Additionally, along with the lane graph, our model also provides BEV object instances, thus offering a comprehensive understanding of the local traffic scene. Our extensive experimental comparisons with powerful baselines demonstrate the superior performance of the proposed method,  in both lane graph and object detection tasks. 

%and validate the proposed lane graph formulation and the metrics. 

\paragraph{Acknowledgements:} The authors gratefully acknowledge support by Toyota Motors Europe (TME).

{\small
\bibliographystyle{ieee_fullname}
\bibliography{egbib}

\begin{thebibliography}{10}\itemsep=-1pt

\bibitem{DBLP:conf/cvpr/AcunaLKF18}
David Acuna, Huan Ling, Amlan Kar, and Sanja Fidler.
\newblock Efficient interactive annotation of segmentation datasets with
  polygon-rnn++.
\newblock In {\em 2018 {IEEE} Conference on Computer Vision and Pattern
  Recognition, {CVPR} 2018, Salt Lake City, UT, USA, June 18-22, 2018}, pages
  859--868. {IEEE} Computer Society, 2018.

\bibitem{DBLP:conf/rss/BansalKO19}
Mayank Bansal, Alex Krizhevsky, and Abhijit~S. Ogale.
\newblock Chauffeurnet: Learning to drive by imitating the best and
  synthesizing the worst.
\newblock In Antonio Bicchi, Hadas Kress{-}Gazit, and Seth Hutchinson, editors,
  {\em Robotics: Science and Systems XV, University of Freiburg, Freiburg im
  Breisgau, Germany, June 22-26, 2019}, 2019.

\bibitem{batra2019improved}
Anil Batra, Suriya Singh, Guan Pang, Saikat Basu, CV Jawahar, and Manohar
  Paluri.
\newblock Improved road connectivity by joint learning of orientation and
  segmentation.
\newblock In {\em Proceedings of the IEEE/CVF Conference on Computer Vision and
  Pattern Recognition}, pages 10385--10393, 2019.

\bibitem{nuscenes2019}
Holger Caesar, Varun Bankiti, Alex~H. Lang, Sourabh Vora, Venice~Erin Liong,
  Qiang Xu, Anush Krishnan, Yu Pan, Giancarlo Baldan, and Oscar Beijbom.
\newblock nuscenes: A multimodal dataset for autonomous driving.
\newblock {\em arXiv preprint arXiv:1903.11027}, 2019.

\bibitem{can2020understanding}
Yigit~Baran Can, Alexander Liniger, Ozan Unal, Danda Paudel, and Luc Van~Gool.
\newblock Understanding bird's-eye view semantic hd-maps using an onboard
  monocular camera.
\newblock {\em arXiv preprint arXiv:2012.03040}, 2020.

\bibitem{DBLP:journals/corr/abs-2012-03040}
Yigit~Baran Can, Alexander Liniger, Ozan Unal, Danda~Pani Paudel, and Luc~Van
  Gool.
\newblock Understanding bird's-eye view semantic hd-maps using an onboard
  monocular camera.
\newblock {\em CoRR}, abs/2012.03040, 2020.

\bibitem{DBLP:conf/eccv/CarionMSUKZ20}
Nicolas Carion, Francisco Massa, Gabriel Synnaeve, Nicolas Usunier, Alexander
  Kirillov, and Sergey Zagoruyko.
\newblock End-to-end object detection with transformers.
\newblock In Andrea Vedaldi, Horst Bischof, Thomas Brox, and Jan{-}Michael
  Frahm, editors, {\em Computer Vision - {ECCV} 2020 - 16th European
  Conference, Glasgow, UK, August 23-28, 2020, Proceedings, Part {I}}, volume
  12346 of {\em Lecture Notes in Computer Science}, pages 213--229. Springer,
  2020.

\bibitem{casas2021mp3}
Sergio Casas, Abbas Sadat, and Raquel Urtasun.
\newblock Mp3: A unified model to map, perceive, predict and plan.
\newblock {\em arXiv preprint arXiv:2101.06806}, 2021.

\bibitem{chen2020learning}
Dian Chen, Brady Zhou, Vladlen Koltun, and Philipp Kr{\"a}henb{\"u}hl.
\newblock Learning by cheating.
\newblock In {\em Conference on Robot Learning (CoRL)}, 2020.

\bibitem{DBLP:conf/eccv/ChenZPSA18}
Liang{-}Chieh Chen, Yukun Zhu, George Papandreou, Florian Schroff, and Hartwig
  Adam.
\newblock Encoder-decoder with atrous separable convolution for semantic image
  segmentation.
\newblock In Vittorio Ferrari, Martial Hebert, Cristian Sminchisescu, and Yair
  Weiss, editors, {\em Computer Vision - {ECCV} 2018 - 15th European
  Conference, Munich, Germany, September 8-14, 2018, Proceedings, Part {VII}},
  volume 11211 of {\em Lecture Notes in Computer Science}, pages 833--851.
  Springer, 2018.

\bibitem{Cordts2016Cityscapes}
Marius Cordts, Mohamed Omran, Sebastian Ramos, Timo Rehfeld, Markus Enzweiler,
  Rodrigo Benenson, Uwe Franke, Stefan Roth, and Bernt Schiele.
\newblock The cityscapes dataset for semantic urban scene understanding.
\newblock In {\em Proc. of the IEEE Conference on Computer Vision and Pattern
  Recognition (CVPR)}, 2016.

\bibitem{cui2019multimodal}
Henggang Cui, Vladan Radosavljevic, Fang-Chieh Chou, Tsung-Han Lin, Thi Nguyen,
  Tzu-Kuo Huang, Jeff Schneider, and Nemanja Djuric.
\newblock Multimodal trajectory predictions for autonomous driving using deep
  convolutional networks.
\newblock In {\em International Conference on Robotics and Automation (ICRA)},
  pages 2090--2096, 2019.

\bibitem{auclair1999survey}
A Fortier, Djemel Ziou, Costas Armenakis, and S Wang.
\newblock Survey of work on road extraction in aerial and satellite images.
\newblock {\em Center for Topographic Information Geomatics, Ontario, Canada.
  Technical Report}, 241(3), 1999.

\bibitem{garg2021semantics}
Sourav Garg, Niko S{\"u}nderhauf, Feras Dayoub, Douglas Morrison, Akansel
  Cosgun, Gustavo Carneiro, Qi Wu, Tat-Jun Chin, Ian Reid, Stephen Gould,
  et~al.
\newblock Semantics for robotic mapping, perception and interaction: A survey.
\newblock {\em arXiv preprint arXiv:2101.00443}, 2021.

\bibitem{he2017}
Kaiming He, Georgia Gkioxari, Piotr Dollár, and Ross Girshick.
\newblock Mask r-cnn, 2017.
\newblock cite arxiv:1703.06870Comment: open source; appendix on more results.

\bibitem{hendy2020fishing}
Noureldin Hendy, Cooper Sloan, Feng Tian, Pengfei Duan, Nick Charchut, Yuesong
  Xie, Chuang Wang, and James Philbin.
\newblock Fishing net: Future inference of semantic heatmaps in grids.
\newblock {\em arXiv preprint arXiv:2006.09917}, 2020.

\bibitem{DBLP:conf/cvpr/HomayounfarMLU18}
Namdar Homayounfar, Wei{-}Chiu Ma, Shrinidhi~Kowshika Lakshmikanth, and Raquel
  Urtasun.
\newblock Hierarchical recurrent attention networks for structured online maps.
\newblock In {\em 2018 {IEEE} Conference on Computer Vision and Pattern
  Recognition, {CVPR} 2018, Salt Lake City, UT, USA, June 18-22, 2018}, pages
  3417--3426. {IEEE} Computer Society, 2018.

\bibitem{homayounfar2018hierarchical}
Namdar Homayounfar, Wei-Chiu Ma, Shrinidhi~Kowshika Lakshmikanth, and Raquel
  Urtasun.
\newblock Hierarchical recurrent attention networks for structured online maps.
\newblock In {\em Proceedings of the IEEE Conference on Computer Vision and
  Pattern Recognition}, pages 3417--3426, 2018.

\bibitem{homayounfar2019dagmapper}
Namdar Homayounfar, Wei-Chiu Ma, Justin Liang, Xinyu Wu, Jack Fan, and Raquel
  Urtasun.
\newblock Dagmapper: Learning to map by discovering lane topology.
\newblock In {\em Proceedings of the IEEE/CVF International Conference on
  Computer Vision}, pages 2911--2920, 2019.

\bibitem{hong2019rules}
Joey Hong, Benjamin Sapp, and James Philbin.
\newblock Rules of the road: Predicting driving behavior with a convolutional
  model of semantic interactions.
\newblock In {\em Proceedings of the IEEE Conference on Computer Vision and
  Pattern Recognition}, pages 8454--8462, 2019.

\bibitem{hou2019learning}
Yuenan Hou, Zheng Ma, Chunxiao Liu, and Chen~Change Loy.
\newblock Learning lightweight lane detection cnns by self attention
  distillation.
\newblock In {\em Proceedings of the IEEE/CVF International Conference on
  Computer Vision}, pages 1013--1021, 2019.

\bibitem{jaritz20202d}
Maximilian Jaritz.
\newblock {\em 2D-3D scene understanding for autonomous driving}.
\newblock PhD thesis, PSL Research University, 2020.

\bibitem{DBLP:journals/corr/abs-2002-06604}
YeongMin Ko, Jiwon Jun, Donghwuy Ko, and Moongu Jeon.
\newblock Key points estimation and point instance segmentation approach for
  lane detection.
\newblock {\em CoRR}, abs/2002.06604, 2020.

\bibitem{DBLP:journals/corr/abs-2012-12160}
Justin Liang, Namdar Homayounfar, Wei{-}Chiu Ma, Shenlong Wang, and Raquel
  Urtasun.
\newblock Convolutional recurrent network for road boundary extraction.
\newblock {\em CoRR}, abs/2012.12160, 2020.

\bibitem{liang2019convolutional}
Justin Liang, Namdar Homayounfar, Wei-Chiu Ma, Shenlong Wang, and Raquel
  Urtasun.
\newblock Convolutional recurrent network for road boundary extraction.
\newblock In {\em Proceedings of the IEEE/CVF Conference on Computer Vision and
  Pattern Recognition}, pages 9512--9521, 2019.

\bibitem{liang2018end}
Justin Liang and Raquel Urtasun.
\newblock End-to-end deep structured models for drawing crosswalks.
\newblock In {\em Proceedings of the European Conference on Computer Vision
  (ECCV)}, pages 396--412, 2018.

\bibitem{DBLP:journals/pami/LinGGHD20}
Tsung{-}Yi Lin, Priya Goyal, Ross~B. Girshick, Kaiming He, and Piotr
  Doll{\'{a}}r.
\newblock Focal loss for dense object detection.
\newblock {\em {IEEE} Trans. Pattern Anal. Mach. Intell.}, 42(2):318--327,
  2020.

\bibitem{DBLP:conf/nips/LiuLMSFSY18}
Rosanne Liu, Joel Lehman, Piero Molino, Felipe~Petroski Such, Eric Frank, Alex
  Sergeev, and Jason Yosinski.
\newblock An intriguing failing of convolutional neural networks and the
  coordconv solution.
\newblock In Samy Bengio, Hanna~M. Wallach, Hugo Larochelle, Kristen Grauman,
  Nicol{\`{o}} Cesa{-}Bianchi, and Roman Garnett, editors, {\em Advances in
  Neural Information Processing Systems 31: Annual Conference on Neural
  Information Processing Systems 2018, NeurIPS 2018, 3-8 December 2018,
  Montr{\'{e}}al, Canada}, pages 9628--9639, 2018.

\bibitem{DBLP:journals/ral/LuMD19}
Chenyang Lu, Marinus Jacobus~Gerardus van~de Molengraft, and Gijs Dubbelman.
\newblock Monocular semantic occupancy grid mapping with convolutional
  variational encoder-decoder networks.
\newblock {\em {IEEE} Robotics Autom. Lett.}, 4(2):445--452, 2019.

\bibitem{ma2019exploiting}
Wei-Chiu Ma, Ignacio Tartavull, Ioan~Andrei B{\^a}rsan, Shenlong Wang, Min Bai,
  Gellert Mattyus, Namdar Homayounfar, Shrinidhi~Kowshika Lakshmikanth, Andrei
  Pokrovsky, and Raquel Urtasun.
\newblock Exploiting sparse semantic hd maps for self-driving vehicle
  localization.
\newblock In {\em 2019 IEEE/RSJ International Conference on Intelligent Robots
  and Systems (IROS)}, pages 5304--5311. IEEE, 2019.

\bibitem{DBLP:journals/corr/abs-2002-08394}
Kaustubh Mani, Swapnil Daga, Shubhika Garg, N.~Sai Shankar, Krishna~Murthy
  Jatavallabhula, and K.~Madhava Krishna.
\newblock Monolayout: Amodal scene layout from a single image.
\newblock {\em CoRR}, abs/2002.08394, 2020.

\bibitem{DBLP:journals/ral/PanSLAZ20}
Bowen Pan, Jiankai Sun, Ho~Yin~Tiga Leung, Alex Andonian, and Bolei Zhou.
\newblock Cross-view semantic segmentation for sensing surroundings.
\newblock {\em {IEEE} Robotics Autom. Lett.}, 5(3):4867--4873, 2020.

\bibitem{pan2020cross}
Bowen Pan, Jiankai Sun, Ho~Yin~Tiga Leung, Alex Andonian, and Bolei Zhou.
\newblock Cross-view semantic segmentation for sensing surroundings.
\newblock {\em IEEE Robotics and Automation Letters}, 5(3):4867--4873, 2020.

\bibitem{paz2021tridentnet}
David Paz, Hengyuan Zhang, and Henrik~I Christensen.
\newblock Tridentnet: A conditional generative model for dynamic trajectory
  generation.
\newblock {\em arXiv preprint arXiv:2101.06374}, 2021.

\bibitem{philion2020lift}
Jonah Philion and Sanja Fidler.
\newblock Lift, splat, shoot: Encoding images from arbitrary camera rigs by
  implicitly unprojecting to 3d.
\newblock In {\em Proceedings of the European Conference on Computer Vision},
  2020.

\bibitem{ravi2018real}
B Ravi~Kiran, Luis Roldao, Benat Irastorza, Renzo Verastegui, Sebastian Suss,
  Senthil Yogamani, Victor Talpaert, Alexandre Lepoutre, and Guillaume Trehard.
\newblock Real-time dynamic object detection for autonomous driving using prior
  3d-maps.
\newblock In {\em Proceedings of the European Conference on Computer Vision
  (ECCV) Workshops}, pages 0--0, 2018.

\bibitem{rella2021decoder}
Edoardo~Mello Rella, Jan-Nico Zaech, Alexander Liniger, and Luc Van~Gool.
\newblock Decoder fusion rnn: Context and interaction aware decoders for
  trajectory prediction.
\newblock {\em arXiv preprint arXiv:2108.05814}, 2021.

\bibitem{DBLP:journals/pami/RenHG017}
Shaoqing Ren, Kaiming He, Ross~B. Girshick, and Jian Sun.
\newblock Faster {R-CNN:} towards real-time object detection with region
  proposal networks.
\newblock {\em {IEEE} Trans. Pattern Anal. Mach. Intell.}, 39(6):1137--1149,
  2017.

\bibitem{richards1999remote}
John~Alan Richards and JA Richards.
\newblock {\em Remote sensing digital image analysis}, volume~3.
\newblock Springer, 1999.

\bibitem{DBLP:conf/cvpr/RoddickC20}
Thomas Roddick and Roberto Cipolla.
\newblock Predicting semantic map representations from images using pyramid
  occupancy networks.
\newblock In {\em 2020 {IEEE/CVF} Conference on Computer Vision and Pattern
  Recognition, {CVPR} 2020, Seattle, WA, USA, June 13-19, 2020}, pages
  11135--11144. {IEEE}, 2020.

\bibitem{seif2016autonomous}
Heiko~G Seif and Xiaolong Hu.
\newblock Autonomous driving in the icity—hd maps as a key challenge of the
  automotive industry.
\newblock {\em Engineering}, 2(2):159--162, 2016.

\bibitem{sun2019leveraging}
Tao Sun, Zonglin Di, Pengyu Che, Chun Liu, and Yin Wang.
\newblock Leveraging crowdsourced gps data for road extraction from aerial
  imagery.
\newblock In {\em Proceedings of the IEEE/CVF Conference on Computer Vision and
  Pattern Recognition}, pages 7509--7518, 2019.

\bibitem{ventura2018iterative}
Carles Ventura, Jordi Pont-Tuset, Sergi Caelles, Kevis-Kokitsi Maninis, and Luc
  Van~Gool.
\newblock Iterative deep learning for road topology extraction.
\newblock {\em arXiv preprint arXiv:1808.09814}, 2018.

\bibitem{yang2018hdnet}
Bin Yang, Ming Liang, and Raquel Urtasun.
\newblock Hdnet: Exploiting hd maps for 3d object detection.
\newblock In {\em Conference on Robot Learning}, pages 146--155. PMLR, 2018.

\bibitem{zaech2020action}
Jan-Nico Zaech, Dengxin Dai, Alexander Liniger, and Luc Van~Gool.
\newblock Action sequence predictions of vehicles in urban environments using
  map and social context.
\newblock In {\em International Conference on Intelligent Robots and Systems
  (IROS)}, 2020.

\end{thebibliography}
}

\newpage
\twocolumn[\begin{center}
      \Large\textbf{Structured Bird's-Eye-View Traffic Scene Understanding from Onboard Images - Supplementary Material}\\
  
   \end{center}
   ]

\section{Summary}

In the supplementary material we present our data augmentation technique, which uses view synthesis to generate more diverse training examples. We give more insight into our proposed connectivity metric, and finally show more visual and quantitative results.

\section{Training}

During training of the method, we apply artificial depthwise motion as a data augmentation. With the flat world assumption, it is possible to calculate the new pixel location of a real world point if the ego vehicle moves $\beta$ in the depth direction. Let the original pixel row and column coordinates be $(m_0, n_0)$ and the new coordinates be $(m_1, n_1)$. Then $n_0 = (n_1-d_x)fC/(fC - m_1\beta + d_y\beta) + d_x$ and $m_0 = (m_1-d_y)fC/(fC - m_1\beta + d_y\beta) + d_x$ where $f$ is the focal length, $C$ is the camera height and $(d_x,d_y)$ are the frame center coordinates. We resample the original image and translate the ground truth (GT) object and centerline points by $\beta$.

\section{Connectivity metric}

The mathematical definition of our connectivity metric and how true positives, false positives and false negatives are defined, is given in the main text. Here we would like to summarize the definition in words. Therefore, let us first give the mathematical definition:
Let the estimated binary incidence matrix be $E$ and the GT incidence matrix be $I$. Let $M(i)$ be the index of the target that the $i$th estimation is matched to and $S(n)$ be the set of indices of estimations that are matched to target $n$. A positive entry $E_{ij}$ is a true positive if $(M(i) == M(j)) \;|\; (I(M(i), M(j)) == 1)$, and a false positive otherwise. A false negative is a positive entry $I_{m,n}$ where $\nexists \;(i,j) : ((i \in S(m))\; \& \;(j \in S(n))\; \& \;(E_{i,j}==1))$. 

In words, if two estimated centerlines are associated, there are two possible ways for this association to be true:
\begin{itemize}
    \item Both estimations are matched with the same target.
    \item The distinct targets that the two estimated centerlines are matched to are, indeed, associated according to the GT incidence matrix.
\end{itemize}

A miss, or a false negative, is present if there is a positive entry $(m,n)$ in the GT incidence matrix $I$ and at least one of the following conditions hold: 
\begin{itemize}
    \item No estimation was matched with target $m$.
    \item No estimation was matched with target $n$.
    \item Among all pairs of estimated centerlines $(i,j)$ where $i$ is matched with target $m$ and $j$ with $n$, there is no pair whose association estimate is positive.
\end{itemize}

\section{Lane graph results}
In the main paper we presented visual results for the lane graph, here we show further examples (see Fig.~\ref{fig:lane-visual}) and explain how we visualized the lane graphs.

\subsection{Lane graph merging method}
Whenever the network is estimating a full lane graph (this excludes PINET, which does not estimate a graph) we merge the predicted lane graph for visualization. The merging works by post processing the Bezier control points and the incidence matrix estimation in the following way:
% In lane graph visual results, merged lane graph is shown whenever applicable (PINET does not estimate associations so we can not apply merging). Given the Bezier control points and the incidence matrix estimation, we apply the following post-processing step for producing visual results:
\begin{itemize}
    \item Extract all junction points where at least 2 centerlines meet.
    \item For all the junctions, get the start point locations of outgoing lines and end points of incoming lines.
    \item Concatenate all the junction points and take the mean, producing one (x,y) pair for each junction.
    \item Replace the start points of outgoing and endpoints of incoming lines with their respective junction locations.
\end{itemize}

\begin{figure*}
    \centering
    \includegraphics[width=\linewidth]{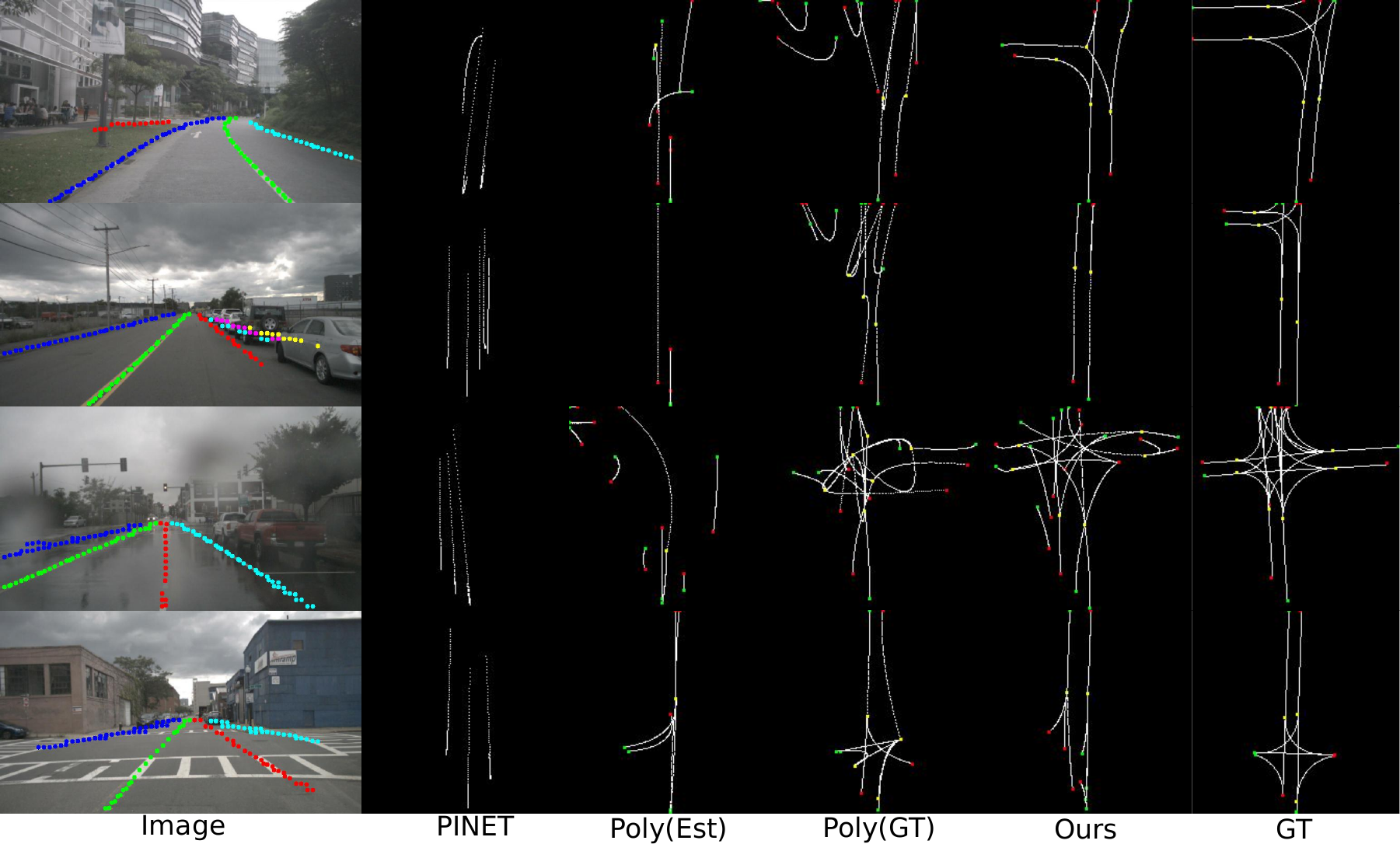}
    \caption{Sample centerline estimates. PINET boundary estimations are shown on the image. Our method produces the best lane graph representation. The detailed statistical results for the scenes in this figure are given.}
    \label{fig:lane-visual}
\end{figure*}

Note that, this process does not change the underlying directed graph but it is useful for visualization. It is possible to formulate more advanced post-processing steps, for example to compute the exact junction point locations one could consider the confidence of centerlines. However, this is beyond the scope of this work.

\subsection{Visual results}
First of all, we would like to emphasize that the results are shown for the whole target region of interest, whether it is occluded or not. We observed that sometimes, the methods can estimate lane graph structure in the occluded regions as well. Moreover, due to difficulty in establishing the occluded regions precisely, we have opted for presenting the results in whole field-of-view. Therefore, the results should be interpreted taking this into consideration.
As stated in the main text and shown in Fig.~6, we have complied the statistical results for each method on each given image for each metric. Below, we present an extended version of Fig.~6, which includes four (instead of two) traffic scenes. Additionally to the visual results of the four traffic scenes we give the quantitative results in Tab.~\ref{tab:imagewise_compare1}-\ref{tab:imagewise_compare4}, and discuss the results. 

% SCENE 1

\begin{table}[h]
% \scriptsize{
\begin{center}
\tabcolsep=0.08cm
\begin{tabular}{ |c|c|c|c|c|c|c| }

\hline
Method & M-Pre & M-Rec & Detect & C-Pre & C-Rec & C-IOU  \\
\hline

PINET & 49& 50& 20&-& -& - \\ 

Poly(Est) & 37.9&  33.2& 60.0& 0.0& 0.0&  0.0 \\

Ours& 60.0& 53.4&  60& 75.0& 60.0&  50.0\\
\hline
Poly(GT) & 54.5& 53.5&  70.0&  66.7&  44.4& 36.4  \\
\hline
\end{tabular}
\end{center}
% }
\vspace{-1em}
\caption{Scene 1 Results}
\label{tab:imagewise_compare1}
\end{table}

In Scene 1, it can be seen that all methods manage to detect the straight lanes more or less accurately. However, only our method can detect the left turn. Moreover, we see that Poly(GT) produce inaccurate estimations in that region. Yet, because our method estimates the turn in a slightly wrong distance, both our method and Poly(GT) suffer similarly in the precision-recall metric. The proposed connectivity metric, however, clearly favors our estimation which is also backed by visual inspection.

\begin{figure*}
    \centering
    \includegraphics[width=\linewidth]{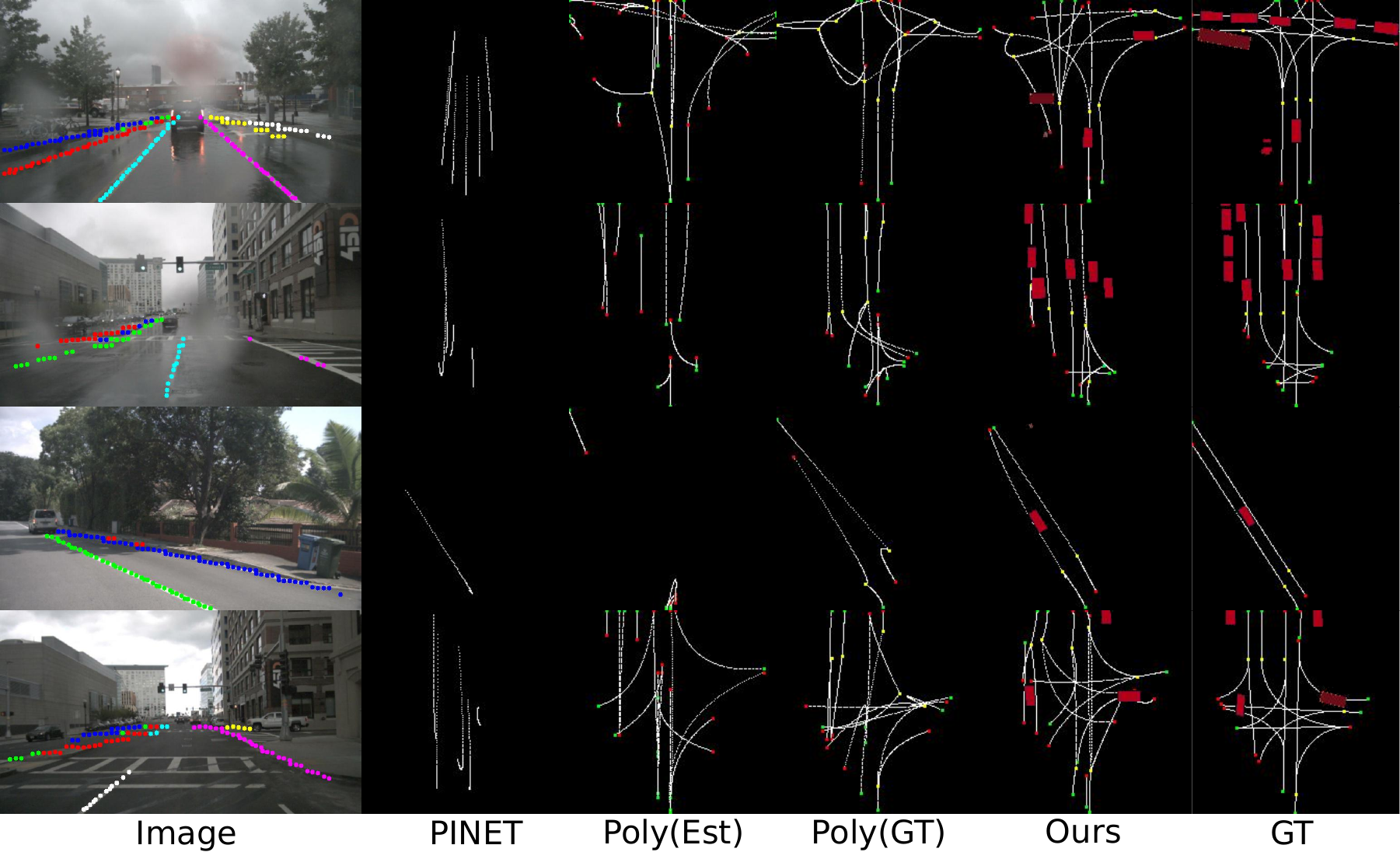}
    \caption{Some additional examples with object estimations also shown for our method. It can be seen that our method produces the best graphs.}
    \label{fig:extra}
\end{figure*}

% SCENE 2

\begin{table}[h]
% \scriptsize{
\begin{center}
\tabcolsep=0.08cm
\begin{tabular}{ |c|c|c|c|c|c|c| }

\hline
Method & M-Pre & M-Rec & Detect & C-Pre & C-Rec & C-IOU  \\
\hline

PINET & 38& 39&20& -& - &- \\ 

Poly(Est) &  70.4&62.5&20.0&0.0& 0.0& 0.0 \\

Ours&  84.5& 77.7& 50.0&  60.0& 33.3& 27.3 \\
\hline
Poly(GT) & 53.4& 66.9& 70.0& 50.0&25.0&20.0 \\
\hline
\end{tabular}
\end{center}
% }
\vspace{-1em}
\caption{Scene 2 Results}
\label{tab:imagewise_compare2}
\end{table}

In Scene 2, Poly(GT) produces many false lines that are matched with straight road segments. This causes a decrease in precision-recall. Our method misses the left turn completely, which decreases the detection score, but it faithfully represent the straight lanes. It should be noted that Poly(Est) performs better than Poly(GT) in precision-recall in this image. Note that PINET can not handle the parked cars and generates several centerlines in the carpark area.

% SCENE 3

\begin{table}[h]
% \scriptsize{
\begin{center}
\tabcolsep=0.08cm
\begin{tabular}{ |c|c|c|c|c|c|c| }

\hline
Method & M-Pre & M-Rec & Detect & C-Pre & C-Rec & C-IOU  \\
\hline

PINET & 64&75&8& -&-&- \\ 

Poly(Est) & 32.9&  30.7&26.1& 100&4.4&4.4 \\

Ours& 48.4& 47.6& 74.0& 82.6&67.9& 59.4 \\
\hline
Poly(GT) &  62.2&  62.4& 65.0&   48.1&48.1& 31.7 \\
\hline
\end{tabular}
\end{center}
% }
\vspace{-1em}
\caption{Scene 3 Results}
\label{tab:imagewise_compare3}
\end{table}

Scene 3 shows a complicated road network and the left and right turns are barely visible. However, our method manages to produce a good estimate. The small differences in the exact location of the lines results, however, in lower precision-recall than Poly(GT). PINET only finds the straight lines as expected while Poly(Est) detects some true initial points that are the beginnings of the turns but the Polygon-RNN head fails to produce the lanes. Again, we see that the connectivity metric demonstrates the superiority of the proposed method.

% SCENE 4

\begin{table}[h]
% \scriptsize{
\begin{center}
\tabcolsep=0.08cm
\begin{tabular}{ |c|c|c|c|c|c|c| }

\hline
Method & M-Pre & M-Rec & Detect & C-Pre & C-Rec & C-IOU  \\
\hline

PINET& 39 &46&30 &-&-&-\\ 

Poly(Est) &  80.2&  81.7 & 30 & 100& 60.0 & 60.0\\

Ours &  54.9 & 52.7 &60.0 & 100 & 66.7& 66.7\\
\hline
Poly(GT) & 64.4 & 69.6& 90& 55.6& 83.3 & 50.0 \\
\hline
\end{tabular}
\end{center}
% }
\vspace{-1em}
\caption{Scene 4 Results}
\label{tab:imagewise_compare4}
\end{table}

In Scene 4, most of the crossroads is not visible. PINET mistakenly estimates 3 lanes (4 lane boundaries) while the rightmost one is actually a carpark area. Poly(Est) detects the lines but estimates them to be close to each other and in the same direction. Poly(GT) estimate suffers from faulty association where the rightmost lane is distorted due to a recalculated junction point location. Our method produces the initial part of the turns but fails to estimate the whole crossroads. In this scene Poly(Est) produces the best results except for the detection score. This is due to the fact that is it misses the whole right part of the image, but produces reasonable estimates of the left part.

\subsection{Additional visual results}
In Fig ~\ref{fig:extra}, we present some additional results with object estimations included. 

In the first scene, the GT shows many cars travelling in the horizontal direction but upon inspection of the image, we observe that part of image is not clear. Thus our method's object estimates are reasonable. The road network estimate of our method is vastly superior to all other baselines including Poly(GT). 

In the second image, our method produces all 4 lanes faithfully. Poly(GT) also produces decent estimates. Our method missed the cars on the right side of the scene, but we see that the rain drop is making that region non-visible. 

Scene 3 provides a relatively easy task for all methods. PINET, unsurprisingly, produces an accurate estimate but Poly(Est) failed to produce the lanes. While Poly(GT) manages to somewhat estimate the direction of the lanes, it produces faulty structures in the bottom part of the FOV. Our method produces very accurate estimations for both centerlines and the car.

The last scene demonstrates a very complex crossroads scene. All 4 directions of the crossroads are visible in the image. This results in Poly(Est) producing good estimates, capturing the essence of the lane graph. The Poly(GT) estimate is denser than Poly(Est) and covers a larger area of the true lane graph but it fails to handle the junctions properly. Our method produces a better lane graph estimate but fails especially in the left side of the image. Our object estimates are accurate, with the exception of the truck which is labelled as car.

% {\small
% \bibliographystyle{ieee_fullname}
% \bibliography{egbib}
% }

\end{document}